# Attention-Driven Lightweight Model for Pigmented Skin Lesion Detection

Mingzhe Hu[a], Xiaofeng Yang[b*]

[a]Department of Computer Science and Informatics, Emory University, Atlanta, GA, 30322, USA; [b]Department of Radiation Oncology and Winship Cancer Institute, Emory University, Atlanta, GA, 30322, USA

[*]Corresponding to xiaofeng.yang@emory.edu

This study presents a lightweight pipeline for skin lesion detection, addressing the challenges posed by imbalanced class distribution and subtle or atypical appearances of some lesions. The pipeline is built around a lightweight model that leverages ghosted features and the DFC attention mechanism to reduce computational complexity while maintaining high performance. The model was trained on the HAM10000 dataset, which includes various types of skin lesions. To address the class imbalance in the dataset, the synthetic minority over-sampling technique and various image augmentation techniques were used. The model also incorporates a knowledge-based loss weighting technique, which assigns different weights to the loss function at the class level and the instance level, helping the model focus on minority classes and challenging samples. This technique involves assigning different weights to the loss function on two levels – the class level and the instance level. By applying appropriate loss weights, the model pays more attention to the minority classes and challenging samples, thus improving its ability to correctly detect and classify different skin lesions. The model achieved an accuracy of 92.4%, a precision of 84.2%, a recall of 86.9%, a f1-score of 85.4% with particularly strong performance in identifying Benign Keratosis-like lesions (BKL) and Nevus (NV). Despite its superior performance, the model's computational cost is considerably lower than some models with less accuracy, making it an optimal solution for real-world applications where both accuracy and efficiency are essential.

## 1. INTRODUCTION

Skin lesions are a common type of disease that can be caused by various factors such as infection, allergies, and excessive cell proliferation. They often manifest as abnormal changes in the surface morphology of the skin, such as changes in color, abnormal growth, and changes in texture. Although skin lesions themselves can be easily determined based on external changes in the skin's appearance, accurately and rapidly identifying the type of skin lesion and whether it is malignant is often difficult. Most types of skin lesions are benign and harmless, but skin cancer remains one of the most common types of cancer[1], with millions of new cases diagnosed worldwide each year. Malignant skin cancers, such as melanoma, carry a high risk of spreading to other parts of the body and endangering the patient's life[2]. Rapid diagnosis and treatment can greatly alleviate the suffering of skin disease patients, reduce the likelihood of death caused by the spread of skin cancer, and minimize the trauma caused by treatment. Currently, the diagnosis of skin cancer primarily relies on dermatologists observing the details of the patient's skin at the affected site with the naked eye or a dermascope and making a diagnosis based on certain morphological or textural features. These two methods rely on the clinician's clinical experience, resulting in strong subjectivity and variability. Invasive skin pathology biopsy can further improve diagnostic accuracy, but it is expensive and the waiting time for test results is long, affecting the patient's prognosis. For economically and medically underserved areas, it is quite challenging to have access to professional dermatologists, dermascope, and biopsy equipment. Therefore, developing a rapid, accurate, cost-effective, and deployable automated skin disease diagnosis system in resource-limited environments is of significant importance.

The rapid development of machine learning techniques, especially deep learning, has greatly improved the automated medical imaging systems, leading to unprecedented performance in the domain of lesion/organ segmentation[3, 4], disease detection[5, 6], and medical image synthesis[7-9].

However, despite the state-of-the-art models pushing the performance boundaries forward, they have become increasingly large and cumbersome, demanding high computational power from the computing platforms. This has made it challenging for many theoretically powerful models to be practically and effectively applied. In such circumstances, rather than pursuing incremental improvements in predictive performance, there is a shift towards seeking a lightweight and computationally efficient model for skin lesion detection, enabling the following application scenarios: 1) Effective telemedicine in resource-limited environments; 2) Increased accessibility and affordability of skin lesion detection systems; 3) Deployment on devices with high energy efficiency requirements, such as personal wearable devices; 4) All computations are performed on the end device to avoid concerns regarding the privacy breach that may arise from uploading personal information to central servers.

Many researchers nowadays are attempting to reduce the parameters and latency of models while striving to maintain optimal performance through various approaches. For example, knowledge distillation[10] transfers the knowledge learned from a more powerful, larger model to a compact model. Pruning methods reduce redundant neuron parameters[11]. Low-rank decomposition[12] decreases the number of model parameters through matrix factorization. Additionally, more efficient and compact models have been proposed, such as MobileNet[13], ShuffleNet[14], EfficientNet[15], and SqueezeNet[16]. In our pipeline for skin lesion detection, we adopted a ghosted structure[17, 18] to remove redundant feature map information from the model while achieving a compute-efficient attention mechanism that captures global features. We also addressed the imbalanced class distribution of skin lesions by using a novel loss function. Compared to other competing state-of-the-art lightweight models, our pipeline achieves better classification performance while maintaining its compactness. The contribution of our research are three folds: 1) We developed a compact skin lesion detection model specifically designed for deployment in resource-limited environments; 2) We investigated and explored the potential benefits of removing redundant feature maps in the context of skin lesion detection. By eliminating unnecessary information, we were able to enhance the computational efficiency of the model without sacrificing much performance; 3) We validated the effectiveness of an efficient attention mechanism design for compact convolutional neural network (CNN) models. This design choice proved to be highly beneficial in capturing important global features relevant to skin lesion detection, contributing to improved performance while maintaining a compact model size.

## 2. METHODS

2.1 HAM10000 Dataset

HAM10000 is an open-source dataset that encompasses the majority of pigmented skin lesion types and is often used as a benchmark dataset for skin lesion classification and detection tasks. The HAM10000 dataset provides a diverse range of skin lesion types, including Melanocytic nevi (NV), Melanoma (MEL), Basal cell carcinoma (BCC), Actinic keratoses (AKIEC), Benign keratosis-like lesions (BKL), Dermatofibroma (DF), and Vascular lesions (VASC). It is worth noting that the class distribution within the HAM10000 dataset is imbalanced, with NV being the most common class representing benign moles and VASC being the rarest class, encompassing various skin abnormalities involving blood vessels.

This dataset consists of a total of 10,015 dermatoscope images, and each image's label has been verified by dermatologists to ensure accuracy. Considering the moderate size of the HAM10000 dataset, we performed a random split to divide the data into three parts: 80% for training, 10% for validation, and the remaining 10% for testing. Figure 1 shows sample images of the seven types of skin lesions included in the dataset.

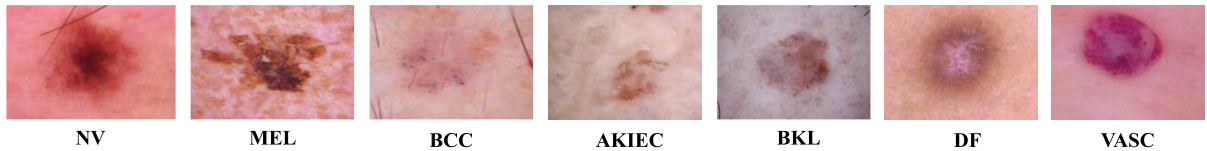

    NV         MEL        BCC       AKIEC     BKL        DF       VASC

**Figure 1.** A selection of sample images representing various skin lesion types included in the dataset. From left to right, the lesions are as follows: Melanocytic nevi (NV), Melanoma (MEL), Basal cell carcinoma (BCC), Actinic keratoses (AKIEC), Benign keratosis-like lesions (BKL), Dermatofibroma (DF), and Vascular lesions (VASC).

In our study, we employ a combination of Synthetic Minority Over-sampling Technique (SMOTE) and various image augmentation techniques to enrich our dataset. SMOTE is used to address class imbalance by generating new, synthetic examples in the minority class, thereby enhancing the diversity of our training data. Alongside SMOTE, we also implement random translation and flipping of images as part of our data augmentation strategy. These techniques help our model learn to recognize skin lesions from different angles and positions, thus improving its generalization capability. Furthermore, to ensure consistency in input data and to optimize computational efficiency, all images are resized to a standard dimension of 224x224 pixels. Lastly, we apply normalization to the images.

2.2 Less Computation with Ghost Features

For CNNs, convolution layers involve computationally intensive operations[19]. These layers convolve the input features with a set of filters, generating the output features. The number of floating-point operations (FLOPs) required for convolution is determined by both the size and quantity of the filters. While reducing the filter size can decrease computational requirements, it also has notable impacts on the model's performance. Reducing the filter size results in smaller receptive fields, which means the model captures less contextual information. Additionally, using smaller filters leads to more frequent downsampling of spatial information, resulting in a loss of spatial details and a failure to capture global features effectively. Simply reducing the size of filters can limit the model's capacity to learn complex patterns and compromise its performance in downstream tasks.

On the other hand, the number of filters in the model may be redundant[20] for the current task. In other words, only a subset of filters (referred to as primary filters) is necessary for generating the intrinsic feature maps, while the remaining "ghost" features can be derived from the intrinsic features through inexpensive linear operations[17]. More specifically, for the conventional convolutional layers, given the input feature maps $I$, the output feature maps $O = I * W$, where $W$ denotes the convolutional kernel and $*$ denotes the convolution operation. The bias term is omitted for simplicity. As for the ghosted CNN structures, the intrinsic feature maps $O' = I * W'$, where $W'$ is the primary convolutional kernel whose number of filters is less than W thus consume less computation resources. Let's denote the number of intrinsic feature maps as $x$ and the number of conventional feature maps as $z$, where $x < z$. To address the potential reduced expressive and discriminative power caused by the reduced number of feature maps in the intermediate layers of ghosted models, we adopt a strategy to maintain the same number of feature maps as in conventional models while still keeping computational resources low. In this approach, we generate the ghosted feature maps from the intrinsic feature maps using a simple linear operation denoted as $f$.

Given the $i_{th}$ intrinsic feature map $o'_i \in O'$, we can generated $s$ ghosted feature map $o_{i,j}$ using function $f_{i,j}$ for $i = 1, ..., x, j = 1, ..., s$. $o_{i,j} = f_{i,j}(o'_i)$. The last linear function $f_{i,s}$ is an identity function that keeps the intrinsic feature maps. Figure 2 visualizes the process of generating the ghosted feature maps from an intrinsic feature map via linear operations. The linear operation is implemented by 3x3 linear kernels (depth-wise convolution[21]). By using these simple linear functions, we obtain $x \cdot s = z$ feature maps. This approach ensures the ghosted model retains the expressive and discriminative power similar to conventional models, while benefiting from the computational efficiency provided by the reduced number of filters and the use of simple linear operations to generate ghosted feature maps.

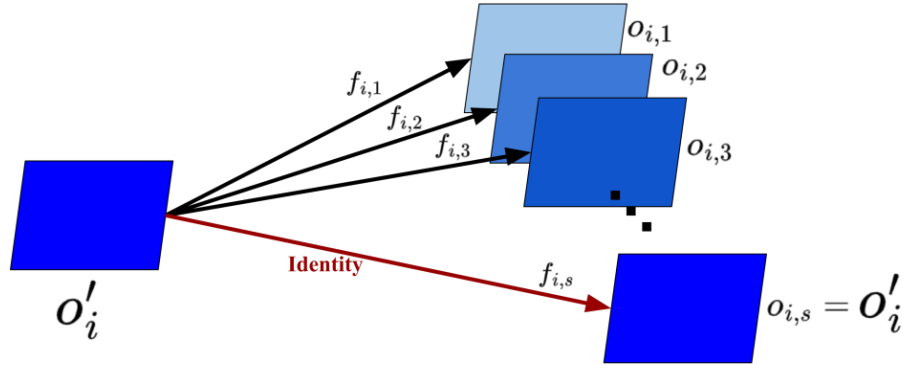

**Figure 2.** The process of generating ghosted feature maps from an intrinsic feature map using linear operations. These simple linear functions efficiently produce ghosted feature maps while retaining the intrinsic model's expressive and discriminative power. Importantly, this method maintains the same number of feature maps output as conventional convolutional filters.

2.3 Efficient Attention Mechanism

Self-attention mechanism was first proposed in natural language processing to obtain contextually enhanced representations for each word[22]. Researchers recognized that the self-attention mechanism's ability to capture global dependencies could also be beneficial for computer vision tasks. The Vision Transformer (ViT)[23] is one of the pioneering models in this regard, showcasing how self-attention can be effectively applied to image classification tasks. Additionally, the Swin Transformer[24] introduces hierarchical attention mechanisms for improved efficiency.

Self-attention mechanism requires storing attention scores for all pairs of elements in the sequence causing a quadratic computational complexity. The high computational complexity can be prohibitive for mobile devices with limited computation power and the corresponding long latency makes it undesirable for real-time applications or interactive user experience.

To enable the application of self-attention mechanism for lightweight models, sparse approximation of the attention mechanism is desirable to achieve faster and memory-efficient computations. One such attention with hardware-friendly operations is the decoupled fully connected (DFC) attention[18]. There are two main contributions to the sparsity of the approximation of the attention mechanism in this approach. Firstly, we replace the self-attention operations with fully-connected (FC) layers to calculate the feature maps. This modification helps reduce the computational complexity and memory requirements compared to the traditional dense self-attention mechanism. Secondly, we decouple the token aggregation process into the vertical and horizontal directions (of the image). By doing so, we can capture long-range dependencies and interactions between tokens more efficiently while maintaining sparsity in the attention computation. More specifically, in our approach, we begin with feature maps $I \in R^{H \times W \times C}$, where $H, W$, and $C$ represent the feature height, feature width, and number of channels, respectively. We split these feature maps into $H \times W$ tokens ($x_i, i = 1, 2, \ldots, H \times W$), where each token contains C values, representing a local region of the input data.

To generate the attention map, the tokenized feature maps are fed through two fully-connected (FC) layers sequentially. The first FC layer operates horizontally, capturing interactions and dependencies between tokens along the width dimension ($W$), and the second FC layer operates vertically, capturing interactions and dependencies between tokens along the height dimension ($H$). The transformation of the first FC layer can be written as: $A'_h = \sum_{h'=1}^{H} \theta^H_{h,h'w} \odot x_{h'w}, h = 1, \ldots, H, w = 1, \ldots, W$, and the second transformation can be written as $A_{hw} = \sum_{w'=1}^{W} \theta^W_{w,hw'} \odot x'_h, h = 1, \ldots, H, w = 1, \ldots, W$. Here $\odot$ is the pointwise multiplication, $\theta^H$ are the weights of the vertical transformation, $\theta^W$ are the weights of the horizontal transformation, $A$ is the derived attention map. By developing the sparse representation, we efficiently process the feature maps and generate the attention maps with reduced computational complexity from $\mathcal{O}(H^2W^2)^2$ to $\mathcal{O}(H^2W + HW^2)$.

2.4 Formulation of the Lightweight Model

In our lightweight model, we adopt the concept of "bottlenecks" from ResNet[25]. Figure 3 illustrates a building block of our model. The input feature map undergoes two separate paths: the ghost module path, which generates ghost features, and the DFC attention path. To combine information from these two paths, we perform a pointwise multiplication between the output feature map of the ghost path and the output feature map of the attention path. The Sigmoid function scales the attention map values to the range (0, 1). The attention map acts as a spatial mask, highlighting the most important regions of the ghost feature maps and suppressing less relevant regions. This leads to more effective and informative representations for the given task. The enhanced feature maps are then sent to the second module, where the number of channels is reduced to match the shortcut path. We use our efficient bottlenecks to replace the bottlenecks of MobileNet-V3 while maintaining the main architecture the same. This change allows our model to retain the advantages of MobileNet-V3[26] while improving efficiency and performance, thanks to the newly designed bottleneck.

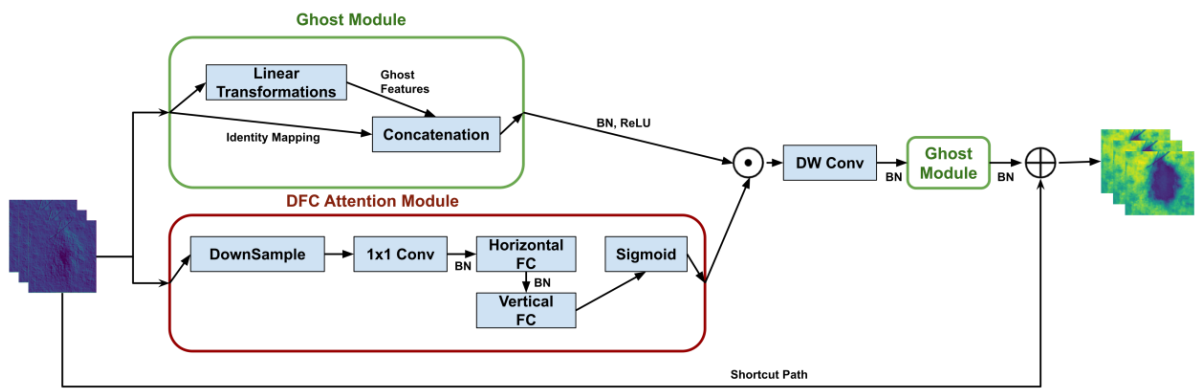

**Figure 3.** The diagram illustrates our lightweight model, featuring two Ghost modules. The first Ghost module expands the feature map channels, while the second one reduces them. Additionally, we incorporate the DFC attention module to efficiently capture global feature relationships. The abbreviation "DWConv" denotes Depthwise Convolution, and "BN" represents batch normalization. To further optimize computation power, the feature maps are downsampled before calculating the attention maps.

2.5 Knowledge-based Loss Weighting

The HAM10000 dataset indeed poses several challenges for accurate skin lesion detection due to its characteristics. Firstly, the dataset suffers from an imbalanced class distribution, where different types of skin lesions have significantly different numbers of instances as shown in figure 4. This class imbalance may cause the model to overly focus on the majority class and result in poor performance on the minority class. Secondly, certain samples in the dataset can be more challenging to detect accurately, even for human experts. Some lesions may have subtle or atypical appearances, making them difficult to classify correctly. To address these challenges and improve the model's performance, we propose adopting knowledge-based loss weighting. This technique involves assigning different weights to the loss function on two levels – the class level and the instance level. By applying appropriate loss weights, we can help the model pay more attention to the minority classes and challenging samples, thus improving its ability to correctly detect and classify different skin lesions.

On the class level, we compensate for the underrepresented classes with more weights and assign less weight to the majority classes. The weight is defined as $\frac{N}{N_i}, i \in \{1,2,\ldots,7\}$, where $N$ is the total number of instances of all classes, and $N_i$ is the total number of instances in class $i$.

On the sample level, we can adjust the loss for each sample based on the difficulty assessment by human experts. In the case of the HAM10000 dataset, along with the images, the dataset also provides notes from human experts. These notes contain information about the diagnostic methods used for each sample, which are categorized as "expert consensus," "serial imaging showed no change," "confocal microscopy," and "Histopatholy." We hypothesize that the difficulty of diagnosis increases from left to right in the mentioned diagnostic methods. Thus, for cases that require more expensive and accurate diagnostic methods, the model needs to pay more attention and learn from them. As a result, we can assign more weights to the corresponding loss for these challenging samples. By incorporating sample-level loss weighting, the model can effectively focus on learning from difficult cases and allocate its learning resources more efficiently. This approach helps improve the model's performance and accuracy, especially in handling samples that require more accurate diagnostic methods, leading to enhanced skin lesion detection capabilities on the HAM10000 dataset.

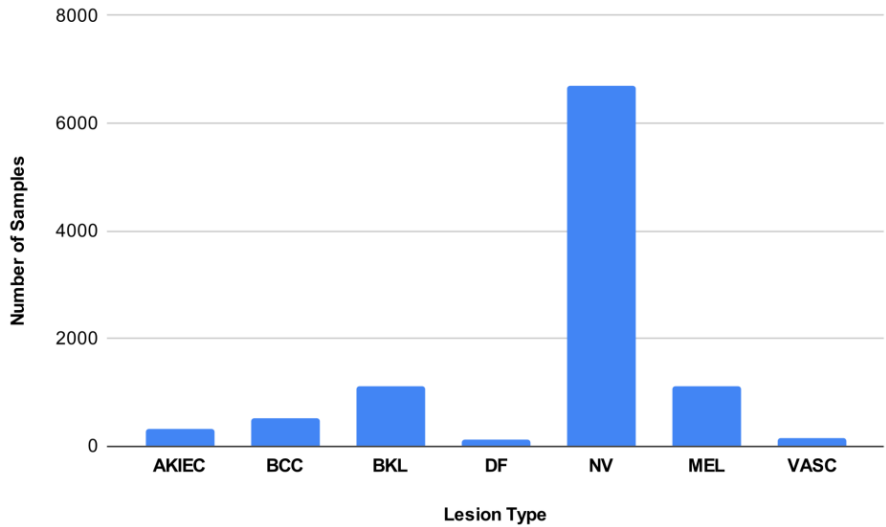

**Figure 4.** The class distribution of skin lesions is imbalanced with melanocytic nevi (NV) as the majority class and actinic keratoses (AKIEC) as the minority class.

## 3. RESULTS

In our experimental setup, our attention-driven lightweight model, grounded on ghost structures, was trained over 100 epochs with early stopping on the validation set to mitigate overfitting. Training was performed using mini-batches of 64 images, utilizing Stochastic Gradient Descent (SGD) with momentum as the optimizer and a starting learning rate of 0.0001. Performance of the model was assessed using the testing subset based on top-1 accuracy, precision, recall, and F1-score, with these metrics presented as macro averages for model comparison. The evaluation included a class-specific performance of our model in table 1 and a comparison against multiple models to highlight the distinctive advantages of our design and methodology in table 2.

**Table 1.** Classification Performance of the Attention-Driven Lightweight Model for Each Skin Lesion Category

| Lesion Type | Precison | Recall | F1-score | Number of Images |
|---|---|---|---|---|
| AKIEC | 0.85 | 0.85 | 0.85 | 33 |
| BCC | 0.71 | 0.88 | 0.79 | 51 |
| BKL | 0.96 | 0.88 | 0.92 | 110 |
| DF | 0.83 | 0.83 | 0.83 | 12 |
| MEL | 0.78 | 0.91 | 0.84 | 111 |
| NV | 0.98 | 0.94 | 0.96 | 671 |
| VASC | 0.79 | 0.79 | 0.79 | 14 |
| Macro Average | 0.84 | 0.87 | 0.85 | 1002 |
| Weighted Average | 0.93 | 0.92 | 0.93 | 1002 |

The classification report indicates that our attention-driven lightweight model has demonstrated strong performance across various classes of skin lesions. Overall, the model achieved an accuracy of 92.4%, indicating a high degree of correctness in classification. The model was particularly successful in identifying Benign Keratosis-like lesions (BKL) and Nevus (NV), achieving high F1-scores of 0.919 and 0.961, respectively, showcasing its ability to balance both precision and recall in these classes.

When we look at the macro-averages, which consider the balance across all classes, we note an F1-score of 0.854, suggesting that our model has delivered a robust performance, even when taking into account potential class imbalances in the dataset. Furthermore, the model achieved an impressive Area Under the Receiver Operating Characteristic (AUROC) score of 89.7%, further underlining its strong discriminative power between different classes.

Some classes such as Basal Cell Carcinoma (BCC) and Melanoma (MEL) presented more of a challenge for the model. Despite this, the model achieved high recall rates for these classes, with 0.882 and 0.910, respectively, which is critical in a medical context where missing a positive case can have serious consequences. Figure 5 shows the confusion matrix of the multi-class skin lesions classification.

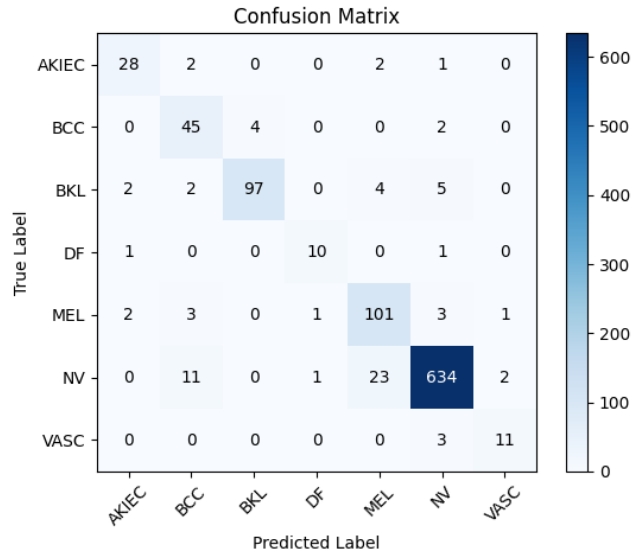

**Figure 5.** Confusion matrix of the attention-driven lightweight model highlighting the performance across different skin lesion classes

These results show that our attention-driven lightweight model has great potential in the automated detection and classification of pigmented skin lesions, offering an efficient tool for early skin cancer detection.

**Table 2.** Comparative Evaluation of Our Model Against Other Compact Lightweight Models in Terms of Macro-Averaged Performance Metrics and Computational Efficiency

| Model | Accuracy | Precision | Recall | F1-score | FLOPs(M) |
|---|---|---|---|---|---|
| MobileNet | 79.3% | 76.9% | 80.0% | 78.4% | 218.8 |
| MobileNetV3 | 83.9% | 83.3% | 83.0% | 83.1% | 83.3 |
| GhostNet | 83.7% | 83.4% | 83.8% | 83.6% | 53.65 |
| EfficientNet B0 | 83.0% | 82.1% | 83.4% | 82.7% | 148.4 |
| ShuffleNet | 84.8% | 83.1% | 80.7% | 81.8% | 55.6 |
| **Ours** | **92.4%** | **84.2%** | **86.9%** | **85.4%** | **63.6** |

Table 2 presents a comparative analysis of our attention-driven lightweight model against other compact, lightweight models in terms of macro-averaged performance metrics and computational efficiency. The models compared include MobileNet, MobileNetV3, GhostNet, EfficientNet B0, and ShuffleNet. Our model shows a notable advantage, achieving the highest accuracy of 92.4% and an F1-score of 85.4%, outperforming the other models by a significant margin. The comparison of performance is visualized as figure 6.

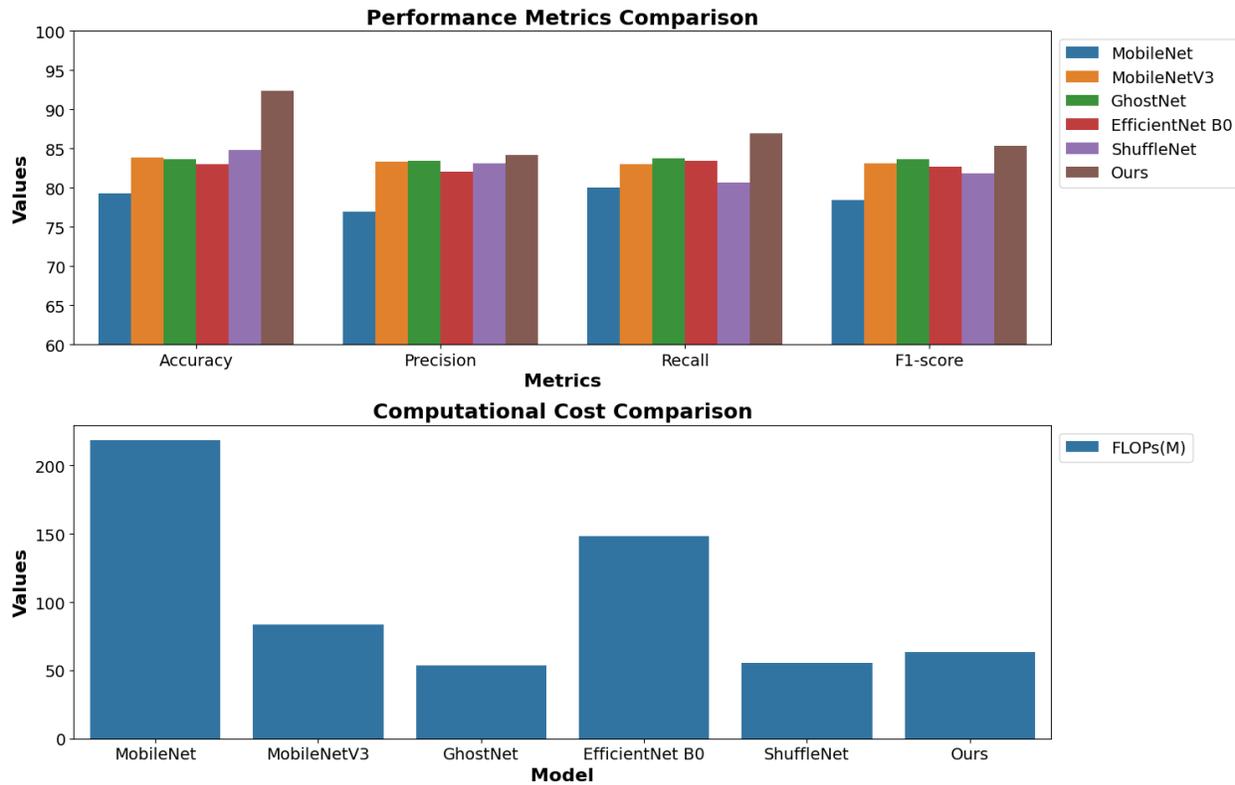

**Figure 6.** Visualization of comparative analysis across different compact lightweight models highlighting performance metrics and computational efficiency

In terms of precision and recall, our model reports 84.2% and 86.9% respectively, again surpassing the comparative models. This indicates our model's balanced capability to correctly identify positive cases (precision) and to detect actual positives from all relevant instances (recall).

Perhaps one of the most impressive features of our model lies in its computational efficiency. Despite its superior performance, the model's computational cost measured in FLOPs (Floating Point Operations Per Second) is 63.6M, which is considerably lower than some models with less accuracy, such as MobileNet and EfficientNet B0. Only GhostNet and ShuffleNet have lower computational costs, but they fall short in the performance metrics.

In summary, these results highlight our model's ability to deliver top-tier performance while maintaining a relatively low computational cost, making it an optimal solution for real-world applications where both accuracy and efficiency are essential.

## 4. CONCLUSION AND DISCUSSION

Our study proposes an efficient and effective pipeline for skin lesion detection, featuring a powerful yet lightweight model. To reduce computational complexity (FLOPs), we replace standard features with ghosted features to minimize redundancy in the feature maps. Furthermore, we enhance the model's ability to capture global features using the DFC attention mechanism, serving as a sparse replacement for the self-attention mechanism. By decoupling the full attention into vertical and horizontal attention, we further reduce computation costs.

Comparing our model to other state-of-the-art lightweight models, our approach achieves the best overall performance while maintaining low computational costs. Despite the success of our pipeline, there are limitations and potential areas for improvement in the future. One limitation is that we adopted the overall structure of MobileNet-V3, and although efficient, further exploration of the overall optimum structure is still warranted.

In the future, we aim to extend this strategy to other medical imaging modalities like MR and CT, which are 3D in nature. However, challenges remain in deriving 3D ghost features and adapting the DFC attention mechanism to 3D space. Additionally, we plan to compare our models with more works and explore the application of our pipelines on real-world mobile devices.

In conclusion, our proposed pipeline demonstrates promising results in skin lesion detection. As we continue to address limitations and explore new possibilities, we envision its potential impact on various medical imaging applications and real-world deployment scenarios.

# REFERENCES

[1] Z. Apalla, A. Lallas, E. Sotiriou *et al.*, "Epidemiological trends in skin cancer," Dermatology practical & conceptual, 7(2), 1 (2017).

[2] C. H. O'Neill, and C. R. Scoggins, "Melanoma," Journal of surgical oncology, 120(5), 873-881 (2019).

[3] S. Pan, C. W. Chang, T. Wang *et al.*, "Abdomen CT multi-organ segmentation using token-based MLP-Mixer," Medical Physics, 50(5), 3027-3038 (2023).

[4] M. Hu, Y. Li, and X. Yang, "Breastsam: A study of segment anything model for breast tumor detection in ultrasound images," arXiv preprint arXiv:2305.12447, (2023).

[5] M. Hu, J. Wang, J. Wynne *et al.*, "A vision-GNN framework for retinopathy classification using optical coherence tomography." 12465, 124650Z.

[6] Y. Li, J. Wang, M. Hu *et al.*, "Prostate Gleason score prediction via MRI using capsule network." 12465, 514-519.

[7] S. Pan, T. Wang, R. L. Qiu *et al.*, "2D medical image synthesis using transformer-based denoising diffusion probabilistic model," Physics in Medicine & Biology, 68(10), 105004 (2023).

[8] S. Pan, E. Abouei, J. Wynne *et al.*, "Synthetic CT Generation from MRI using 3D Transformer-based Denoising Diffusion Model," arXiv preprint arXiv:2305.19467, (2023).

[9] X. Dai, Y. Lei, Y. Fu *et al.*, "Multimodal MRI synthesis using unified generative adversarial networks," Medical physics, 47(12), 6343-6354 (2020).

[10] J. Gou, B. Yu, S. J. Maybank *et al.*, "Knowledge distillation: A survey," International Journal of Computer Vision, 129, 1789-1819 (2021).

[11] T. Hoefler, D. Alistarh, T. Ben-Nun *et al.*, "Sparsity in deep learning: Pruning and growth for efficient inference and training in neural networks," The Journal of Machine Learning Research, 22(1), 10882-11005 (2021).

[12] T. N. Sainath, B. Kingsbury, V. Sindhwani *et al.*, "Low-rank matrix factorization for deep neural network training with high-dimensional output targets." 6655-6659.

[13] A. G. Howard, M. Zhu, B. Chen *et al.*, "Mobilenets: Efficient convolutional neural networks for mobile vision applications," arXiv preprint arXiv:1704.04861, (2017).

[14] X. Zhang, X. Zhou, M. Lin *et al.*, "Shufflenet: An extremely efficient convolutional neural network for mobile devices." 6848-6856.

[15] B. Koonce, and B. Koonce, "EfficientNet," Convolutional Neural Networks with Swift for Tensorflow: Image Recognition and Dataset Categorization, 109-123 (2021).

[16] F. N. Iandola, S. Han, M. W. Moskewicz *et al.*, "SqueezeNet: AlexNet-level accuracy with 50x fewer parameters and< 0.5 MB model size," arXiv preprint arXiv:1602.07360, (2016).

[17] K. Han, Y. Wang, Q. Tian *et al.*, "Ghostnet: More features from cheap operations." 1580-1589.

[18] Y. Tang, K. Han, J. Guo *et al.*, "GhostNetv2: enhance cheap operation with long-range attention," Advances in Neural Information Processing Systems, 35, 9969-9982 (2022).

[19] J. Wu, C. Leng, Y. Wang *et al.*, "Quantized convolutional neural networks for mobile devices." 4820-4828.

[20] Z. Wang, C. Li, and X. Wang, "Convolutional neural network pruning with structural redundancy reduction." 14913-14922.

[21] R. Zhang, F. Zhu, J. Liu *et al.*, "Depth-wise separable convolutions and multi-level pooling for an efficient spatial CNN-based steganalysis," IEEE Transactions on Information Forensics and Security, 15, 1138-1150 (2019).

[22] D. Hu, "An introductory survey on attention mechanisms in NLP problems." 432-448.

[23] A. Vaswani, N. Shazeer, N. Parmar *et al.*, "Attention is all you need," Advances in neural information processing systems, 30, (2017).

[24] Z. Liu, Y. Lin, Y. Cao *et al.*, "Swin transformer: Hierarchical vision transformer using shifted windows." 10012-10022.

[25] K. He, X. Zhang, S. Ren *et al.*, "Deep residual learning for image recognition." 770-778.

[26] A. Howard, M. Sandler, G. Chu *et al.*, "Searching for mobilenetv3." 1314-1324.